\documentclass[sigconf,nonacm]{acmart}

\setcopyright{none}
\acmISBN{}

\usepackage{multirow}
\usepackage{tabularx}

\begin{document}

\title{
Know Your Intent: An Autonomous
Multi-Perspective LLM Agent Framework for DeFi User Transaction Intent Mining
}

\author{Qian'ang Mao}
\email{me@c0mm4nd.com}
\affiliation{%
  \institution{Nanjing University}
  \city{Nanjing}
  \state{Jiangsu}
  \country{China}
}

\author{Yuxuan Zhang}
\email{cassie-ikk@smail.nju.edu.cn}
\affiliation{%
  \institution{Nanjing University}
  \city{Nanjing}
  \state{Jiangsu}
  \country{China}
}

\author{Jiaman Chen}
\email{jiamanchen@smail.nju.edu.cn}
\affiliation{%
  \institution{Nanjing University}
  \city{Nanjing}
  \state{Jiangsu}
  \country{China}
}

\author{Wenjun Zhou}
\email{wzhou4@utk.edu}
\affiliation{%
 \institution{The University of Tennessee, Knoxville}
 \city{Knoxville}
 \state{Tennessee}
 \country{USA}}

\author{Jiaqi Yan}
\email{jiaqiyan@nju.edu.cn}
\affiliation{%
  \institution{Nanjing University}
  \city{Nanjing}
  \state{Jiangsu}
  \country{China}}





\begin{abstract}

As Decentralized Finance (DeFi) develops, understanding user intent behind DeFi transactions is crucial yet challenging due to complex smart contract interactions, multifaceted on-/off-chain factors, and opaque hex logs. Existing methods lack deep semantic insight. To address this, we propose the Transaction Intent Mining (TIM) framework.
TIM leverages a DeFi intent taxonomy built on grounded theory and a multi-agent Large Language Model (LLM) system to robustly infer user intents. A Meta-Level Planner dynamically coordinates domain experts to decompose multiple perspective-specific intent analyses into solvable subtasks. Question Solvers handle the tasks with multi-modal on/off-chain data. While a Cognitive Evaluator mitigates LLM hallucinations and ensures verifiability.
Experiments show that TIM significantly outperforms machine learning models, single LLMs, and single Agent baselines. We also analyze core challenges in intent inference. This work helps provide a more reliable understanding of user motivations in DeFi, offering context-aware explanations for complex blockchain activity.


\end{abstract}




\maketitle

\section{Introduction}

As the core of Web3, decentralized finance (DeFi) has seen a rapid development in recent years \cite{bongini2025crypto}, introducing an open and permissionless financial infrastructure where users can interact with smart contracts through on-chain transactions to perform operations such as depositing, lending, and trading \cite{harvey2021defi}. Behind these on-chain activities lie specific user intents, referring to the purposes users seek to accomplish through single or a series of operations \cite{mao2025defiintent}.


\begin{figure}[ht]
    \raggedright
    \includegraphics[width=\linewidth]{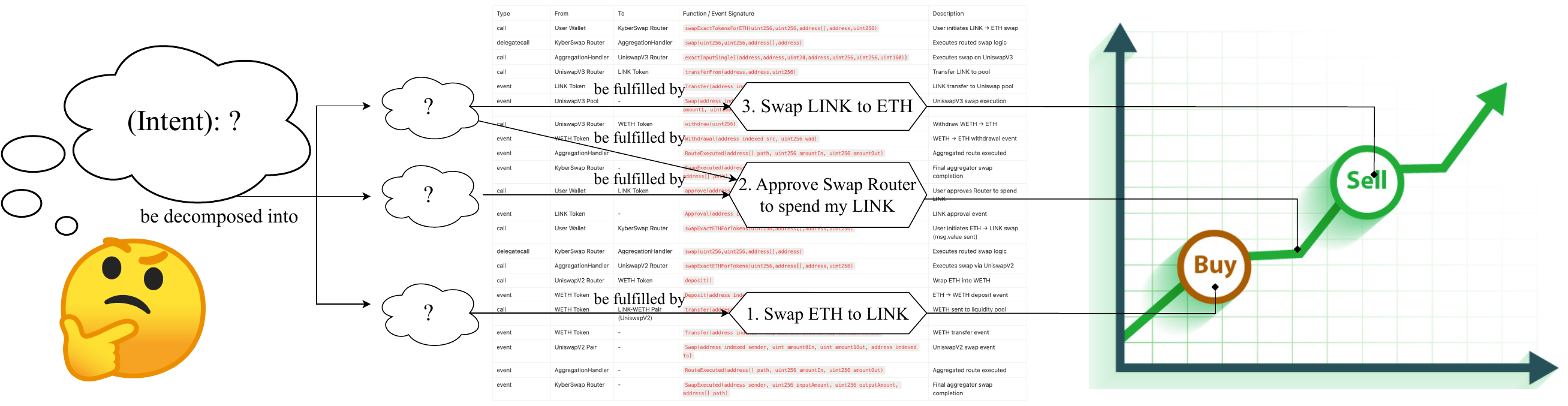}
    \caption{
User's intent on spot trading for profit. Even though it represents one of the simplest and most common intentions, it still requires analyzing the on-chain data behind three related transactions, summarizing the corresponding semantics, and incorporating external market price data in order to understand the purpose of each transaction and thereby infer the underlying intent.
    }
    \label{fig:intent_in_reality}
\end{figure}

The current DeFi ecosystem still faces many challenges, including but not limited to difficulties in matching NFT trading demand \cite{tang2023exploring,kraussl2024non}, frequent threats to capital security \cite{zamani2020security,zhang2024unraveling}, and severe volatility caused by market panic \cite{hairudin2024isotropy,cheraghali2024impact}. In the ``dark forest" environment of DeFi, which is characterized by frequent and drastic fluctuations, a deep understanding of users' transaction intents is crucial \cite{yu2024don,wu2025hunting,si2024understanding}. 
However, the actual process of intent action is extremely complex, as shown in \autoref{fig:intent_in_reality}, which seriously affects the development of the industry\cite{wu2023know}. As emphasized by Blocknative, a Web3 infrastructure enterprise, Web3 remains notoriously complicated, and usability continues to pose a major challenge for the ecosystem\footnote{\url{https://www.blocknative.com/blog/web3-transaction-lifecycle}}.
Therefore, mining the DeFi transaction intents of on-chain users helps reveal the underlying behavioral logic and decision-making motives, thus more accurately depicting user profiles. This also provides more targeted support for almost all on-chain tasks like risk monitoring, product design, and market forecasting. For existing research, incorporating user intent can potentially provide finer-grained semantic information and behavioral explanations for various addresses.


Despite its importance, the need to recognize DeFi transaction intents is underscored. 
Firstly, on-chain transaction records are represented in the form of hexadecimal logs and low-level smart contract calls, making it inherently challenging to decode these records and infer human-readable intent. 
Secondly, on-chain user behavior cannot be fully captured by isolated transaction attributes such as from, to, or value. True intents often hide within complex contract execution logic or across multiple related transactions.
Moreover, DeFi-related on-chain activities do not occur in a vacuum. User behavior on the blockchain is frequently influenced by off-chain factors such as macroeconomic conditions (e.g., interest rate changes, new regulatory policies), mainstream media coverage, social media sentiment, or the opinions of key opinion leaders (KOLs) \cite{gupta2025predicting,luo2025drivers,xing2025toward}. 
Therefore, to accurately infer transaction intents, there is an urgent need for an autonomous and intelligent method that goes beyond single transaction analysis.

To address the challenges, we have designed and implemented the Transaction Intent Mining (TIM) framework, which analyzes transaction intentions through the autonomous behavior of multi-layered agents, which are based on large language models (LLMs).
Our contributions are mainly in the following three aspects:

\begin{itemize}
    \item We designed and implemented a workflow enabling agents to autonomously act and perform DeFi user intent analysis, setting a new standard in the field.
    \item We improved the analysis result by enhancing analytical granularity and reducing the hallucinations of LLMs by integrating on-chain and off-chain data. 
    \item We demonstrated the overall superiority of the framework in the experimental results and discussed the current challenges in DeFi user transaction intent mining tasks by examining the explainable LLM outputs.
\end{itemize}


\section{Related Work}

\subsection{Blockchain Transaction Analysis}
Blockchain transaction analysis has evolved from statistical approaches to sophisticated pattern recognition methodologies, including quantitative metrics \cite{luo2025optimizing}, graph-based fund flow tracking \cite{yousaf2019tracing}, trading strategy identification \cite{qi2023blockchain}, and security monitoring systems \cite{sanjay2023anomaly}. 

Despite these advancements, existing approaches remain predominantly focused on structural and quantitative transaction aspects, exhibiting a significant limitation in the semantic understanding of user intents \cite{wu2023know,song2022blockchain,wu2021analysis}. This limitation is particularly pronounced in DeFi, where complex multi-step transactions and intricate protocol interactions necessitate a deeper semantic comprehension that current approaches fail to provide, thus creating a critical research gap in transaction intent mining.

\subsection{LLM-based Autonomous Agent}
Recent advancements in LLM research have established sophisticated autonomous agent frameworks with enhanced cognitive systems \cite{park2023generative,hu2023chatdb}, domain-specific tool utilization \cite{wang2023voyager,qin2023toolllm}, and collaborative reasoning capabilities \cite{qian2023communicative,dong2024self}. Preliminary applications have already emerged, \citet{li2024cryptotrade} enhance trading decisions through an agent reflection mechanism that analyzes outcomes of past transactions. The capabilities align theoretically with blockchain analysis requirements, offering potential solutions to complex information processing challenges. 

However, a significant research gap exists in applying agents to blockchain transaction intent mining, where semantic interpretation capabilities remain largely unexplored despite the promise for transforming on-chain behavioral analysis beyond the limitations of traditional approaches. 
In our preliminary research, we experimented with existing agent frameworks to address intent mining. We found that frameworks such as AutoGen and LangChain, along with methodologies like ReAct and Conversation, encountered engineering challenges that prevented them from yielding effective results. These frameworks were unable to capture users' complete intents, often stopping at a shallow parsing of basic semantics. Furthermore, when processing on-chain data, these frameworks struggle with the comprehension of low-level data and are hindered by context dependency. This results in critical issues such as factual information loss and exacerbated hallucinations.

\subsection{Intent Mining Methods and Challenges}
Intent mining has progressed from traditional approaches, including rule-based systems and supervised learning models \cite{zhang2021multimodal,huang2018automating} to blockchain-specific methods like MoTS \cite{wu2023know}, yet these techniques struggle with DeFi's dynamic nature and complex behavioral sequences. Emerging LLM-based approaches \cite{bodonhelyi2024user,hong2024dial,zhang2024tinid} offer promising alternatives through advanced language processing capabilities and contextual understanding. 

A critical gap persists as current LLM applications predominantly address scenarios with natural language inputs, these intent recognition tasks \cite{wu2020slotrefine,pham2023misca,qin2021gl} are all trying to learn intent through labeled sentence fragments, which neglects blockchain transaction intent analysis with its unique challenges of complex interactions and patterns on on-chain structured data and off-chain information, which underscores the need for approaches designed specifically for multi-step reflective analysis. 
%
\begin{figure*}[htbp]
  \centering
  \includegraphics[width=1\linewidth]{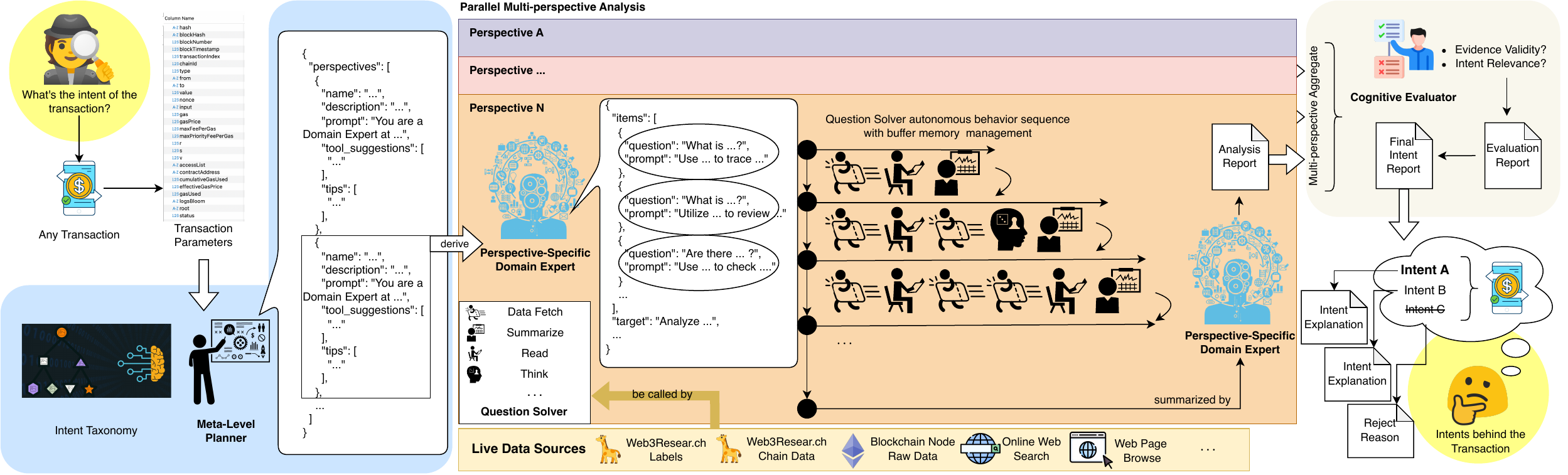}
  \caption{The Transaction Intent Mining (TIM) Framework initiates with a Meta-Level Planner processing transaction parameters and an intent taxonomy to derive perspective-specific domain experts, which then conduct parallel multi-perspective analysis by executing question-solving sequences leveraging live data sources to generate analysis reports, which are aggregated and cognitively evaluated for evidence validity and intent relevance to produce a Final Intent Report detailing validated transaction intents and explanations.
  }
    \label{fig:framework}
\end{figure*}

\section{TIM: A Structurally Deterministic Workflow Framework for Intent Mining}



As shown in \autoref{fig:framework}, the Transaction Intent Mining (TIM) framework structure is a structurally deterministic workflow for solving the intent mining task. This chapter will unfold according to the overall workflow, focusing on the functional design of various agents and systematically analyzing the entire framework around the core challenges faced in actual practice.

\subsection{Problem Formulation}


We formalize the problem of mining transaction intent as an unsupervised multi-label assignment task. Given a transaction $\texttt{tx}$, our goal is to design an analysis system $M$ to analyze the transaction $\texttt{tx}$ and determine the potential set of intents $I_{\texttt{tx}}$ in all intents $I$ from the taxonomy.
In the implementation, we propose a multi-agent intent mining framework, formulated as the analytical system $M = \{A_1, A_2, \dots, A_n; \mathcal{C}\}$, where each $A_i$ is an autonomous LLM-based agent, and $\mathcal{C}$ denotes the coordination mechanism that organically integrates these agents through structured collaboration and interaction. Our research focuses on the design of $A_n$ and $\mathcal{C}$ to enable effective multi-agent cooperation, which enhances the inferred result $I_{\texttt{tx}}$.

\subsection{DeFi Transaction Intent Taxonomy Derivation}

\begin{table*}[htbp]
\centering
\begin{tabular}{p{4cm}p{4cm}p{7cm}}
\toprule
\textbf{Core Category} & \textbf{Axial Coding} & \textbf{Secondary Coding} \\
\midrule
\multirow{12}{*}{\parbox{4cm}{Investment or Speculative Profit-seeking}} & \multirow{4}{*}{\parbox{4cm}{B1 Trading Strategies}} & {A1} Spot Trading Profit \\
\cline{3-3}
 &  & {A2} Leveraged Trading Profit \\
\cline{3-3}
 &  & {A3} Long-term Holding  \\
\cline{3-3}
 &  & {A4} Arbitrage  \\
\cline{2-3}
 & \multirow{3}{*}{\parbox{4cm}{B2 Liquidity Mining and Yield Farming}} & {A5} Provide/Create Liquidity Pool \\
\cline{3-3}
 &  & {A6} Participating in Lending \\
\cline{3-3}
 &  & {A7} Yield Aggregation \\
\cline{2-3}
 & \multirow{3}{*}{\parbox{4cm}{B3 Staking}} & {A8} ETH Liquid Staking \\
\cline{3-3}
 &  & {A9} DeFi Governance Token Staking \\
\cline{3-3}
 &  & {A10} Compound Liquid Staking \\
\cline{2-3}
 & \multirow{2}{*}{\parbox{4cm}{B4 Early Project Participation}} & {A11} Participating in Airdrops \\
\cline{3-3}
 &  & {A12} Participating in Presales/Initial Offerings \\
\cline{1-3}
\multirow{5}{*}{\parbox{4cm}{Personal Risk Control and Management}} & \multirow{3}{*}{\parbox{4cm}{B5 Asset Security Assurance}} & {A13} Using Secure Wallets \\
\cline{3-3}
 &  & {A14} Permission Management \\
\cline{3-3}
 &  & {A15} Purchasing Insurance \\
\cline{2-3}
 & \multirow{2}{*}{\parbox{4cm}{B6 Investment Risk Management}} & {A16} Stop-loss Strategies \\
\cline{3-3}
 &  & {A17} Hedging Strategies \\
\cline{1-3}
\multirow{4}{*}{\parbox{4cm}{Project Participation and Ecosystem Governance}} & \multirow{2}{*}{B7 Direct Governance} & {A18} Voting \\
\cline{3-3}
 &  & {A19} Proposals \\
\cline{2-3}
 & \multirow{2}{*}{B8 Indirect Governance} & {A20} Delegating Voting Rights \\
\cline{3-3}
 &  & {A21} Vulnerability Reporting \\
\bottomrule
\end{tabular}
\caption{DeFi User Intent Core Category Coding Framework}
\label{tab:intent_tax}
\end{table*}


To start, we need a comprehensive DeFi user transaction intent taxonomy to obtain the DeFi transaction intent set $I$ and to guide and carry out the subsequent intent mining work.
In our previous work\cite{mao2025defiintent}, we developed the DeFi transaction intent taxonomy using a Grounded Theory approach, following \citet{corbin1990grounded}. 
Specifically, we drew data from semi-structured interviews with DeFi users of varying experience and extensive textual data from official tutorials and community forums for 51 representative DeFi protocols. 

Our iterative coding process began with open coding of these data sources, through which we identified 110 initial concepts of user intent. We then systematically grouped these concepts during axial coding into 21 broader categories. Finally, through selective coding, we further refined these categories into nine main themes, which we synthesized into three overarching dimensions: Investment or Speculative Profit-seeking, Personal Risk Control and Management, and Project Participation and Ecosystem Governance, all unified by the core theoretical concept of `Risk-reward Trade-off Orientation.' 

We verified theoretical saturation by analyzing additional data from the community and protocols, which confirmed stability and comprehensiveness. 
Final results are shown in the \autoref{tab:intent_tax}.

\subsection{Reflective Multimodal Live Data Retrieval}

Since a single transaction decision can be influenced by numerous factors, such as DeFi protocol details, contract methods, market fluctuations, etc., multimodal live data retrieval is an essential and crucial component for analyzing blockchain transaction intents. However, existing blockchain data research has only analyzed limited scenarios within given datasets, which poses a significant obstacle to practical applications. Therefore, we first implemented live data retrieval with tool use functions, allowing agents to seamlessly link the latest blockchain transactions with the latest internet and blockchain data.

To further enhance the retrieval capabilities, we incorporated a reflective design in the data retrieval process. Specifically, after acquiring data, we use prompts to help understand and critique the data, aiding them in optimizing their retrieval methods in the next execution. For example, they can fetch more data, switch data sources, or further process the data. The reflective design is beneficial for handling complex structured data, such as smart contract function arguments and event logs, in multiple steps, and for uncovering the sources and destinations of funds through complex contexts.

In addition, we also need to address the issue of multimodality: the various types of structured data in blockchain, HTML data from web pages, and text data from agent communications need to be unified. To achieve this, we simplified the blockchain data by JSON formatting and removing content that is not suitable for machine reading, such as Merkle trees, while retaining important readable information like from and to addresses and values; we kept the main body tag text information from HTML; and ultimately, all data was unified into a pure text format between agents.

\subsection{Meta-Level Planner (MP)}


The first to start working is a meta-level planning agent with meta-cognitive capabilities, whose core task is to dynamically plan multi-perspective analysis routes based on transaction context. Each perspective represents a subsequent expert cognitive model, used to decompose actual tasks from specific angles. This design originates from the fact that on-chain DeFi transactions have extremely high heterogeneity and opacity, and a single analysis path often fails to comprehensively cover the complex intents behind the transactions. Traditional static rules or black-box models lack scalability and explainability. Therefore, we need an agent that can dynamically plan analysis paths, adapt to various transaction structures, and possess semantic regulation capabilities.
Based on literature and industry experience, we guide the agent to conduct an in-depth analysis in the following aspects:

\noindent\textbf{Smart Contract Analysis}: This derived agent specializes in analyzing the structure and functionality of blockchain smart contracts. Drawing from program understanding and static analysis theories, it focuses on interpreting contract code, ABI interfaces, and interaction patterns to infer the purpose of user-contract interactions. 

\noindent\textbf{Temporal Context Analysis}: This derived agent implements temporal cognitive modeling methods to analyze the historical background and associated patterns of transactions. By examining user historical behaviors, related transactions, and market environments, it identifies potential behavioral patterns and intent indicators. 

\noindent\textbf{Market Dynamics Analysis}: Derived agent implements multimodal information fusion mechanisms to focus on off-chain market data and macro environments, contextualizing blockchain transactions within broader economic and social backgrounds. 

In addition, due to the autonomy, there may also be derived other perspectives such as protocol background analysis, hacker behavior analysis, etc., depending on the transaction parameters.




\subsection{Perspective-Specific Domain Experts (DE)}


In actual tasks, the analysis of intent is often not a single linear process but requires handling multi-layered, interdependent composite structures. Such structured intents typically involve multiple sub-goals, reasoning steps, or context dependencies, making the analysis process itself complex. Traditional methods often rely on single paths or static features for judgment, making it difficult to effectively stage and decompose these composite structures. Consequently, they face significant limitations when dealing with complex scenarios.

To address this, we designed a mechanism to simulate domain experts based on the ``divide-and-conquer'' strategy: for each concerned perspective, we introduced an expert agent specialized in that field, who would proactively propose multiple sub-questions and derive intelligent agents to answer and analyze around these questions. DE can more logically capture intents by breaking down complex intent analysis problems into a chain of reasoning in question form. In actual implementation, DE is designed to be processed in parallel due to its independence and lack of interference, which improves analysis efficiency. DE agents use data structure components to plan tasks, with plans containing three key elements: objectives, to-do items, and prompts, forming a complete task description. This structured task representation enables agents to clearly understand their responsibilities and execution direction, enhancing the precision and controllability of task execution.

After all issues have been resolved, the original DE will read and check each answer, analyze the user's transaction intent through the reasoning chain, and finally provide an intent analysis report from that perspective.

\subsection{Question Solvers (QS)}

Each question from DE will derive a Question Solver, which will solve the problem through the most granular task execution, adhering to the ``Single Responsibility Principle." This design significantly reduces task complexity, improves execution efficiency, and enhances result quality. Specific actions will be autonomously decided by the QS without human intervention.

In the specific implementation, as shown in \ref{fig:framework}, although QS is completely derived from DE, the main workflow of QS is to perform data retrieval, then decide the subsequent data processing, continue retrieval, or in-depth thinking, or complete the summary based on the reading results of the data in the ReAct paradigm. The result of each question will be brought into the memory of the next QS, helping the latest QS to solve the current problem based on the results of all previous questions.

\subsection{Cognition Evaluator (CE)}

Although the analysis relies on credible external data, the hallucination issue of LLMs still makes it unreliable to depend on the analysis results generated by LLMs. 

To address this issue, we introduced CE before obtaining the final result. It will critically evaluate the analysis reports generated by all DEs from two dimensions: verifiability of facts and relevance to intent. Verifiability of facts refers to whether the statements, information, or reasoning in the report can be validated by objective and traceable evidence. This critique prevents the LLM from generating hallucinations based on unfounded imagination and speculation about user intent. Relevance to intent refers to the degree to which the final results of the report are related to the user's intent. The purpose of this analysis is to filter out the unsubstantiated claims in irrelevant fields from the model's output, preventing these irrelevant parts from affecting the derivation of the final result.

Ultimately, CE evaluates these two dimensions to rank the possible intents mentioned in the DE's analysis report, excluding incorrect intents, and the remaining intents become the system's output $I_{\texttt{tx}}$, which includes the most likely intent as well as multiple possible intents.

\section{Experiments}

\subsection{Experimental Setup}

\noindent\textbf{Experiment Environment:} 
We conduct all experiments using Python on Arch Linux with 8 × RTX 3090 GPUs. More details are in \autoref{sec:detailed_exp_env}. All code, including prompts, is available on GitHub\footnote{\url{https://github.com/c0mm4nd/TIM}}.

\noindent\textbf{Transaction Sources:}
We
created a list of intent-labeled Ethereum DeFi transactions covering all intent categories, as most of the data needs to be retrieved autonomously on-demand, by agents, from the internet or the blockchain.
The list includes transactions involving mainstream protocols (e.g., Uniswap, Aave, and Compound) and tokens (e.g., WETH, WBTC, USDT, and USDC), up to March 2025. The complete list is covered in the \autoref{tab:protocol_call}. We invited multiple experts in the industry to manually label these transactions to determine intents within the intent taxonomy. Experts assigned transaction intents based on their personal experience with online tools. We adopted a multi-round labeling process and combined an expert consensus validation mechanism to ensure the accuracy and consistency.






\noindent\textbf{Evaluation metrics:} We employed recall, precision, and F1-micro to assess the performance.

\noindent\textbf{TIM Setup and Baselines:} 
In the experiment, we default to using Grok-2 as the base model for TIM, with CRIPSE-like style prompts for initializing and guiding agents.
To evaluate the effectiveness of our framework, we selected multiple types of machine learning models (Naive Bayes, SVM, Decision Tree, XGBoost, and CNN), a single LLM, and a single agent as baseline models. Among them, machine learning uses common transaction characteristics in cutting-edge fields such as blockchain transaction fraud detection, which is shown in the \autoref{tab:transaction-features}; the single LLM means that it only relies on the Grok-2 and tries to complete the task through the prompt engineering and basic transaction information; single agent standing on the single LLM, functioning like other ReAct agent frameworks, with all tools and data sources in QS, and multi-round reflectively thinking. In addition, we also selected different base models for performance comparison within the framework. Besides the originally designed Grok-2, the models include Grok-3, Grok-3-Mini, GPT-4.0, GPT-4o-Mini, LLaMA 3.3, and Qwen 3 30B.
Finally, we conducted a comparison using a different prompt engineering framework, LangGPT.


\subsection{Experimental Results}

\begin{table}[ht]
\centering
\begin{tabular}{lccc}
\toprule
\textbf{Method} & \textbf{Recall} & \textbf{Precision} & \textbf{F1-micro} \\
\midrule
Naive Bayes & 0.76 & 0.36 & 0.49 \\
SVM & 0.59 & 0.50 & 0.54 \\
Decision Tree & 0.65 & 0.44 & 0.52 \\
XGBoost & 0.68 & 0.55 & 0.61 \\
CNN+sigmoid & 0.62 & 0.58 & 0.62 \\
\hline
Single LLM & 0.33 & 0.28 & 0.30 \\
Single Agent & 0.38 & 0.45 & 0.41 \\
w/o MP & 0.62 & 0.65 & 0.63 \\
w/o DE & 0.40 & 0.48 & 0.44 \\
w/o QS & 0.32 & 0.30 & 0.31 \\
w/o CE & 0.72 & 0.45 & 0.55 \\
\hline
\textbf{TIM} & 0.78 & \textbf{0.72} & \textbf{0.75} \\
w/ grok-3-mini) & 0.72 & 0.70 & 0.71 \\
w/ grok-3 & \textbf{0.80} & 0.69 & 0.74 \\
w/ gpt-4o-mini & 0.73 & 0.69 & 0.71 \\
w/ gpt-4o & 0.77 & 0.68 & 0.72 \\
w/ qwen3-30b & 0.69 & 0.67 & 0.66 \\
\hline
w/ LangGPT & 0.75 & 0.71 & 0.73 \\
\bottomrule
\end{tabular}
\caption{Comparative and Ablation Study Results}
\label{tab:experiment_result}
\end{table}

\begin{table}[ht]
\centering
\begin{tabular}{cccc}
\toprule
\textbf{Intent Code} & \textbf{Recall} & \textbf{Precision} & \textbf{F1-micro} \\
\midrule
A1  & 1.00 & 0.73 & 0.84 \\
A2  & 0.67 & 0.53 & 0.59 \\
A3  & 0.82 & 0.73 & 0.77 \\
A4  & 0.60 & 0.56 & 0.58 \\
A5  & 0.53 & 0.55 & 0.54 \\ 
A6  & 0.85 & 0.60 & 0.70 \\
A7  & 0.72 & 0.68 & 0.70 \\
A8  & 0.57 & 0.52 & 0.54 \\ 
A9  & 0.88 & 0.90 & 0.89 \\
A10 & 0.56 & 0.72 & 0.29 \\
A11 & 0.30 & 0.28 & 0.29 \\
A12 & 0.55 & 0.60 & 0.57 \\
A13 & 0.92 & 0.80 & 0.86 \\
A14 & 1.00 & 0.52 & 0.68 \\
A15 & 0.82 & 0.74 & 0.78 \\
A16 & 0.53 & 0.43 & 0.47 \\
A17 & 0.42 & 0.45 & 0.43 \\
A18 & 1.00 & 1.00 & 1.00 \\
A19 & 0.67 & 0.33 & 0.44 \\
A20 & 1.00 & 1.00 & 1.00 \\ 
A21 & 0.50 & 0.50 & 0.50 \\
\bottomrule
\end{tabular}
\caption{TIM Performance on Individual Intent Labels }
\label{tab:intent_diff}
\end{table}


To validate our multi-agent approach, we conducted comparative experiments against machine learning models, single LLM, and single-agent setups. Results are shown in \autoref{tab:experiment_result}.


Our analysis shows that machine learning methods, such as Naive Bayes and Decision Tree, achieved moderate recall but suffered from low precision, indicating frequent false positives. These models often misclassified transactions as Spot Trading Profit (A1), which, while common, led to many incorrect predictions. In contrast, subtler intents like Arbitrage (A4) and Leveraged Trading Profit (A2) were frequently overlooked. This suggests that traditional methods struggle to capture the complex, high-dimensional patterns inherent in semantically rich blockchain data.

In contrast, TIM demonstrates significantly better performance on evaluation metrics. Notably, whether using a LLM of standard scale, a more streamlined small-scale model, or a specialized inference-optimized model, TIM consistently maintains a similar and high level of intent mining performance. This fully demonstrates the robustness and effectiveness of the proposed multi-agent collaborative architecture, indicating that the framework's design can effectively integrate the analytical capabilities of models of different scales and reduce over-reliance on the original performance of specific ultra-LLM. This ensures the quality of analysis while providing greater flexibility and potential efficiency advantages for practical deployment.

Through the analysis of the results in \autoref{tab:intent_diff} and a review of the output reports, we found that for intents with names matching method names, such as "Voting" (A18) and "Delegated Voting Rights" (A20), perfect F1-micro scores were achieved; however, for intents such as "Spot Trading Profit" (A1) and "Permission Management" (A14), which are closely tied to basic foundational operations, although they showed high recall rates, they were not sufficiently precise. Conversely, for more complex and ambiguous intents, TIM faces greater challenges; for example, "Participating in Airdrop" (A11), "Compound Liquidity Mining" (A10), and complex strategies such as "Stop-Loss Strategy" (A16) and "Hedging Strategy" (A17) had relatively lower F1-micro scores. This performance difference highlights our advantage in interpreting transactions with clear on-chain footprints, while also indicating the need for improvement in its accuracy when handling intents that are not uniquely defined, involve multiple steps, or depend on external environmental factors. This may require fine-tuning of LLMs to enhance understanding of DeFi domain-specific knowledge.



In the comparison of prompt engineering frameworks, LangGPT performed relatively less effectively than the default CRIPSE style, although the gap was not substantial.

\subsection{Ablation Study}

To further evaluate the effectiveness of our multi-agent approach, we designed additional ablation experiments with two alternative configurations: a single LLM approach (removing both multi-agent collaboration and external tools) and a single-agent approach (removing only multi-agent collaboration while retaining external tools).

As shown in \autoref{tab:experiment_result}, the single LLM approach, which operated without external tools or multi-agent collaboration, yielded the weakest performance across all metrics. This result underscores the limitations of large language models in interpreting complex domains such as blockchain without contextual knowledge or specialized support. Although the introduction of external tools within a single-agent setup led to some improvement, it still fell short compared to traditional machine learning methods and our full framework. These findings confirm the necessity of multi-agent collaboration and demonstrate the effectiveness of our proposed approach in enhancing LLM-based analysis of on-chain transactions.

In addition, we carried out more fine-grained ablation experiments. Specifically, in the w/o MP setting, MP was replaced with three fixed perspective instructions. In the w/o DE setting, the problem was solved directly using perspective instructions, without domain decomposition or final summarization. In the w/o DS setting, DE conducted the analysis directly, without QS processing. Finally, in the w/o CE setting, the results were aggregated directly, without verification. The results, as shown in \autoref{tab:experiment_result}, demonstrate that each module design is essential.




Since the agent retrieves and processes a large amount of external data, context optimization in reflective multimodal live data retrieval becomes a critical component. Without optimization, most of the data would exceed the LLM’s context length. To enable ablation studies, we therefore applied chunk-and-summarize strategies to these lengthy inputs. For comparison, using grok-2 with our proposed data optimization techniques in reflective multimodal live data retrieval (such as formatting and removing machine-generated data like hashes), the average cost of analyzing a single transaction intent can be reduced from around 10\$ to less than 1\$. Notably, by leveraging a mini model, the analysis cost for most transactions can be further reduced to below 0.1\$, which makes it possible for us to package it as an online service.

\subsection{Sensitivity Analysis}

\begin{table}[htbp]
\begin{tabularx}{\linewidth}{p{0.18\linewidth}p{0.18\linewidth}X X X X}
\toprule
\textbf{Creative Value} & \textbf{Executive Value} & \textbf{Recall} & \textbf{Prec.} & \textbf{F1} \\
\midrule
Temp.=0.0 & Temp.=0.0 & 0.56 & 0.53 & 0.54 \\
Temp.=0.0 & Temp.=0.5 & 0.53 & 0.50 & 0.51 \\
Temp.=0.0 & Temp.=0.8 & 0.50 & 0.48 & 0.49 \\
Temp.=0.5 & Temp.=0.0 & \textbf{0.78} & \textbf{0.72} & \textbf{0.75} \\
Temp.=0.5 & Temp.=0.5 & 0.65 & 0.62 & 0.63 \\
Temp.=0.5 & Temp.=0.8 & 0.61 & 0.57 & 0.59 \\
Temp.=0.8 & Temp.=0.0 & 0.62 & 0.58 & 0.60 \\
Temp.=0.8 & Temp.=0.5 & 0.57 & 0.54 & 0.55 \\
Temp.=0.8 & Temp.=0.8 & 0.53 & 0.51 & 0.52 \\
Top-p=1   & Top-p=1     & \textbf{0.78} & \textbf{0.72} & \textbf{0.75} \\
Top-p=1   & Top-p=0.5   & 0.76 & 0.72 & 0.74 \\
Top-p=0.5 & Top-p=1     & 0.75 & 0.71 & 0.73 \\
Top-p=0.5 & Top-p=0.5   & 0.75 & 0.72 & 0.74 \\
\bottomrule
\end{tabularx}
\caption{LLM Hyperparameter Sensitivity Analysis Results}
\label{tab:sensi}
\end{table}

In our sensitivity analysis, we focused on the LLM’s temperature parameter, which controls output randomness: higher values produce more diverse responses, while lower values result in more deterministic ones.
In TIM, we categorized agent tasks into two types: creative tasks (MP and DE) and execution tasks (QS and CE). As shown in \autoref{tab:sensi}, we found that different temperature settings worked optimally for different types of agent roles within our system. 
Similarly, we also made further adjustments to the top-p parameter of the LLM, but its impact on the analysis results was not significant, possibly because the words in the JSON result were limited by the taxonomy and live data, which had already achieved the pruning effect of top-p.

\subsection{Case Study}

\begin{figure*}[h]
    \centering
    \includegraphics[width=\textwidth]{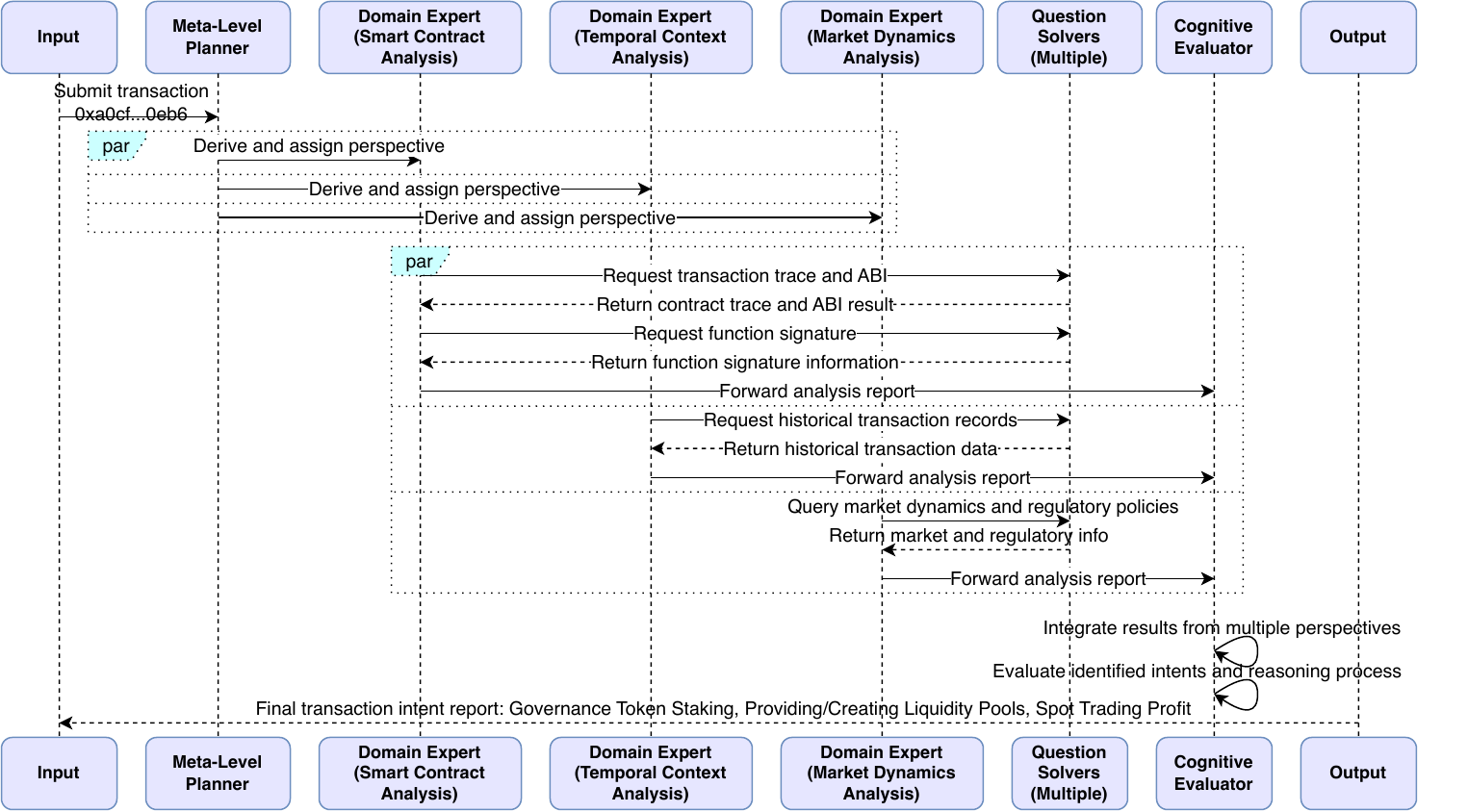}
    \caption{The Intent Mining Sequence Diagram illustrates actions (simplified for showcase) within the Case Study.}
    \label{fig:case-study-workflow}
\end{figure*}




\autoref{fig:case-study-workflow} presents a comprehensive interaction sequence diagram of our framework for mining transaction intent. This case study illustrates a collaborative multi-agent system for blockchain transaction intent analysis. The MP breaks down complex intent mining tasks into three perspectives: Smart Contract Analysis, Temporal Contextual Analysis, and Market Dynamics Analysis. 
This structured decomposition establishes a comprehensive framework that significantly reduces analytical complexity while maintaining thoroughness.

During each domain's analysis phase, DE and QS on each perspective operate concurrently, accessing specific data tools to support their unique analysis, such as parsing smart contract ABI and fetching market information. 
In each concurrent process, the DS first generates a coherent list of questions. For example, in Smart Contract Analysis, it asks which smart contracts are being interacted with, which methods are called, and what the effects of these methods are, etc. The QS then addresses these one by one using external tools and data.
The QS demonstrates dynamic execution capabilities by continuously adjusting its analytical strategy based on emerging evidence. For example, to interpret data from traces and ABIs, the QS will autonomously retrieve function signatures and, as a result, correctly understand transaction semantics.

CE integrates and evaluates intents by analyzing factorial information—such as function signatures in Temporal Context Analysis—which are crucial for determining intent. This process ensures both the correctness and interpretability of the inferred intent before ultimately returning the result to the user.
In this case, the analysis ultimately identified a compound transaction intent involving governance token staking, liquidity provision, and spot trading, where governance token staking was inferred by retrieving fundamental market knowledge, liquidity provision was derived from the yield logic implemented in the token’s smart contract code, and spot trading was determined by combining function signatures with subsequent sell actions.

This workflow represents a progressive reasoning chain construction, which mirrors expert cognitive processes, maintaining flexibility when facing uncertainties and gradually refining hypotheses based on evidence until reaching well-supported conclusions.

\section{Conclusion}
\label{sec:conclusion}


In this paper, we introduced the Transaction Intent Mining (TIM) framework, a novel multi-agent system based on large language models (LLMs) designed to autonomously infer user intents from complex DeFi transactions. The TIM framework employs a self-derived hierarchical agent architecture, including a Meta-Level Planner, Perspective-Specific Domain Experts, Question Solvers, and a Cognition Evaluator. This structure decomposes the intent mining task, integrates multimodal data, and critically evaluates findings to ensure accuracy and mitigate issues like hallucination common in LLMs.
Our empirically derived DeFi intent taxonomy provides a structured basis for analysis. Experimental results demonstrate that TIM significantly outperforms established baselines on all metrics, highlighting the efficacy of its multi-perspective collaborative reasoning and its ability to handle the dynamic and opaque nature of DeFi. This work contributes a more explainable methodology for understanding user intent in DeFi, paving the way for advanced blockchain analytics and user-centric services.




\bibliographystyle{ACM-Reference-Format}
\bibliography{custom}

\appendix

\section*{Appendix}\label{sec:appendix}

\begin{table}[htbp]
\centering
\begin{tabularx}{\linewidth}{X X}
\toprule
\textbf{Token/Protocol/Scene} & \textbf{Function/Event} \\
\midrule
1inch & fillOrderArgs \\
Aave & flashLoan \\
Aave & stake \\
Aave & borrow \\
Aave & repay \\
ArbitrageExecuteStrategy & executeStrategy \\
Azuki & allowlistMint \\
bounty & fundBounty \\
bounty & grantArbiter \\
bounty & makeSubmission \\
Compound & delegateBySig \\
Compound & supply  \\
Compound & withdraw \\
Curve & create \\
dYdX & claimRewards \\
ENS & delegate \\
ENS & delegateBySig \\
ERC20 CRV & approve \\
ERC20 CRV & revoke \\
ERC20 DAI & approve \\
ERC20 DAI & revoke \\
ERC20 USDT & approve \\
ERC20 USDT & revoke \\
ERC20 USDC & approve \\
ERC20 USDC & revoke \\
ERC20 WBTC & approve \\
ERC20 WBTC & revoke \\
ERC20 WETH & approve \\
ERC20 WETH & revoke \\
ERC721 BAYC & - \\
GnosisSafe & createProxy \\
GnosisSafe & execTransaction \\
HarvestFinance & announceStrategyUpdate \\
HarvestFinance & deposit \\
HarvestFinance & withdraw \\
Lido & deposit \\
Lido & deposit\_proxy \\
MakerDAO & frob \\
MakerDAO & join \\
MEVBot & - \\
Multicall3 & aggregate3 \\
NexusMutual & buyCover \\
NotionalFinance & doLeveragedNToken \\
Pendle & swapExactSyForPt \\
Pendle & swapExactPtForSy \\
Rocket & deposit \\
\bottomrule
\end{tabularx}
\caption{The DeFi token, protocol, or scene sources of transactions used in our experiment, filtered by function or event signature, where "-" means sampling any and "revoke" means approve zero amount  }
\label{tab:protocol_call}
\end{table}

\begin{table}[htbp]
\centering
\begin{tabularx}{\linewidth}{X X}
\toprule
\textbf{Token/Protocol/Scene} & \textbf{Function/Event} \\
\midrule
SushiSwap & addLiquidity \\
SushiSwap & removeLiquidity \\
SushiSwap & swapExactETHForTokens \\
SushiSwap & swapExactTokensForETH \\
Swell & deposit \\
Uniswap & addLiquidity \\
Uniswap & claim \\
Uniswap & delegate \\
Uniswap & delegateBySig \\
Uniswap & removeLiquidity \\
Uniswap & swapETHForExactTokens \\
YearnFinance & deposit \\
YearnFinance & withdraw \\
- & vote \\
- & propose \\
\bottomrule
\end{tabularx}
\caption{The DeFi token, protocol, or scene sources of transactions used in our experiment, filtered by function or event signature, where "-" means sampling any and "revoke" means approve zero amount (Cont.d) }
\end{table}

\section{Experiment Environment}\label{sec:detailed_exp_env}

All experiments are conducted on an Arch Linux server equipped with 8 NVIDIA RTX 3090 GPUs, 2TB of memory, and dual AMD EPYC 7763 64-core processors. TIM is developed using the OpenAI SDK framework with the xAI official Grok LLM API, and the compared LLM models use the corresponding official API and OpenRouter API (as a fallback). Baseline machine Learning models are written in Python, utilizing PyTorch and Scikit-learn frameworks.

\section{Data Collection and Preparation}

Our experimental dataset comprises DeFi transaction records from the Ethereum blockchain spanning from project inception to March 2025. The dataset encompasses transactions involving major protocols such as Uniswap, Aave, Compound, and prominent tokens including WETH, WBTC, USDT, USDC, and DAI. We initially identified direct interactions between users and smart contracts, then expanded our search to include transactions triggering the Transfer event within contracts to account for token-based DeFi operations. Through stratified random sampling, we constructed a comprehensive collection of DeFi transaction hashes for experimental analysis, as shown in \autoref{tab:protocol_call}.

To establish ground truth for our intent classification system, we engaged domain experts to annotate each transaction according to our predefined intent taxonomy. The annotation process involved a multi-criteria assessment based on contract type, interaction pattern (direct contract function calls or indirect triggering through token transfers), transaction value, and function call parameters. We implemented a multi-round annotation workflow with an expert consensus validation mechanism to ensure annotation accuracy and consistency. This rigorous process resulted in all transactions being categorized within our predefined intent framework.

\section{Baseline Feature}

To ensure the fairness and scientific rigor of the experimental design, this study draws extensively on research findings from cutting-edge fields such as blockchain transaction fraud detection during the feature selection process. 
From the feature sets commonly used in machine learning methods, we selected the feature indicators shown in \autoref{tab:transaction-features} to describe transaction information in a more comprehensive and domain-specific manner.

\begin{table*}[htbp]
\centering
\begin{tabularx}{\textwidth}{p{0.20\textwidth}p{0.35\textwidth}X}
\toprule
\textbf{Feature} & \textbf{Description} & \textbf{Purpose} \\
\midrule
nonce & A counter for the sending address, indicating the number of transactions it has initiated & Useful for analyzing user behavior patterns, such as distinguishing between regular users and bots (e.g., MEV bots) \\
\hline
transactionIndex & The index of the transaction within the block, representing its order in the block & Helps analyze priority transactions during congestion and understand how miners select transactions \\
\hline
blockNumber & The height of the block containing the transaction, indicating its chronological order & Used to study the temporal distribution of transactions, detect market trends, and identify trading patterns \\
\hline
value & The transaction amount, measured in wei (the smallest Ethereum unit) & Helps identify large-value transactions, analyze whale behavior, and detect market manipulation \\
\hline
gas & The actual amount of gas consumed during the transaction execution & Used to evaluate transaction costs, study gas price dynamics, and optimize smart contract performance \\
\hline
gasPrice & The gas price per unit set at the time of transaction submission, measured in wei & Useful for studying transaction fee fluctuations, gas auction mechanisms, and priority fee strategies \\
\hline
inputLength & The byte length of the transaction input data, usually for contract calls or data storage & Helps analyze interaction patterns with smart contracts and detect abnormal contract invocation behavior \\
\hline
logLength & The number of event logs generated in the transaction receipt & Used to study DeFi transaction patterns such as DEX trading, liquidity provision, and liquidation \\
\hline
traceLength & The number of execution traces generated during the transaction process & Useful for analyzing complex transaction paths such as flash loans, arbitrage, and cross-chain operations \\
\bottomrule
\end{tabularx}
\caption{Transaction Features Used in Supervised Machine Learning Methods}
\label{tab:transaction-features}
\end{table*}

\section{Annotation Process}


To prove the system's effectiveness, we chose to invite experts to perform more annotations to cover most scenarios on-chain rather than using the interview labels from the qualitative research data.
Although the labels are inferred by experts rather than directly reflecting the users’ native intents, they still provide a valid and reliable basis for evaluating the performance of classification systems that are likewise grounded in external analysis.
Transaction intent annotation is a very time-consuming and labor-intensive task, fortunately, our research received significant support from many industry-related organizations. Therefore, we invited experts from the AML (Anti-Money Laundering) department of centralized exchanges (CEX), blockchain fund security practitioners, and blockchain veteran users (with professions including traders, developers, and scholars), totaling 10 people, to help annotate the intents of the 600 sampled transactions.
We did not set a fixed process for the analysis, only requiring that the intent be selected from our previously established intent taxonomy and that online, open tools be used to obtain information. This flexibility for experts is good for capturing diverse approaches but might introduce variability. 
During the process, each transaction was annotated by two independent annotators. The inter-annotator agreement prior to adjudication was Cohen’s kappa = 0.7.
We designated CEX AML experts and blockchain fund security practitioners as the expert adjudication. 
Ultimately, we obtained the final annotated intent results for the 600 transaction data through the expert consensus validation mechanism.




\end{document}